\documentclass[10pt, a4paper]{article}

\usepackage[final]{lrec2026} 
\usepackage{booktabs}
\usepackage{amsmath}

\newcommand{\data}[1]{\csname data:#1\endcsname}
\newcommand{\setvalue}[2]{\expandafter\gdef\csname data:#1\endcsname{#2}}
\setvalue{total_translations}{651}
\setvalue{unique_translations}{334}
\setvalue{num_sources}{13}
\setvalue{num_online_sources}{12}
\setvalue{num_languages}{5}
\setvalue{total_english}{390}
\setvalue{unique_english}{194}
\setvalue{multiplier_english}{5.0}
\setvalue{total_french}{78}
\setvalue{unique_french}{41}
\setvalue{multiplier_french}{2.4}
\setvalue{total_italian}{33}
\setvalue{unique_italian}{17}
\setvalue{multiplier_italian}{2.4}
\setvalue{total_polish}{48}
\setvalue{unique_polish}{29}
\setvalue{multiplier_polish}{4.8}
\setvalue{total_spanish}{102}
\setvalue{unique_spanish}{53}
\setvalue{multiplier_spanish}{2.4}
\setvalue{max_multiplier_lang}{english}
\setvalue{max_multiplier_value}{5.0}
\setvalue{prior_christodouloupoulos_english}{2}
\setvalue{prior_christodouloupoulos_french}{1}
\setvalue{prior_christodouloupoulos_italian}{1}
\setvalue{prior_christodouloupoulos_polish}{1}
\setvalue{prior_christodouloupoulos_spanish}{1}
\setvalue{prior_christodouloupoulos_total}{6}
\setvalue{prior_ebible_english}{24}
\setvalue{prior_ebible_french}{4}
\setvalue{prior_ebible_italian}{2}
\setvalue{prior_ebible_polish}{2}
\setvalue{prior_ebible_spanish}{5}
\setvalue{prior_ebible_total}{37}
\setvalue{prior_mayer_english}{39}
\setvalue{prior_mayer_french}{17}
\setvalue{prior_mayer_italian}{7}
\setvalue{prior_mayer_polish}{6}
\setvalue{prior_mayer_spanish}{22}
\setvalue{prior_mayer_total}{91}
\setvalue{site_bible_audio_english}{3}
\setvalue{site_bible_audio_french}{15}
\setvalue{site_bible_audio_italian}{0}
\setvalue{site_bible_audio_polish}{0}
\setvalue{site_bible_audio_spanish}{1}
\setvalue{site_bible_audio_total}{19}
\setvalue{site_bible_com_english}{75}
\setvalue{site_bible_com_french}{19}
\setvalue{site_bible_com_italian}{7}
\setvalue{site_bible_com_polish}{8}
\setvalue{site_bible_com_spanish}{28}
\setvalue{site_bible_com_total}{137}
\setvalue{site_bible_is_english}{18}
\setvalue{site_bible_is_french}{8}
\setvalue{site_bible_is_italian}{1}
\setvalue{site_bible_is_polish}{2}
\setvalue{site_bible_is_spanish}{15}
\setvalue{site_bible_is_total}{44}
\setvalue{site_biblegateway_com_english}{64}
\setvalue{site_biblegateway_com_french}{4}
\setvalue{site_biblegateway_com_italian}{5}
\setvalue{site_biblegateway_com_polish}{3}
\setvalue{site_biblegateway_com_spanish}{19}
\setvalue{site_biblegateway_com_total}{95}
\setvalue{site_biblehub_com_english}{46}
\setvalue{site_biblehub_com_french}{0}
\setvalue{site_biblehub_com_italian}{0}
\setvalue{site_biblehub_com_polish}{0}
\setvalue{site_biblehub_com_spanish}{0}
\setvalue{site_biblehub_com_total}{46}
\setvalue{site_bibles_org_english}{72}
\setvalue{site_bibles_org_french}{6}
\setvalue{site_bibles_org_italian}{3}
\setvalue{site_bibles_org_polish}{3}
\setvalue{site_bibles_org_spanish}{14}
\setvalue{site_bibles_org_total}{98}
\setvalue{site_biblestudytools_com_english}{37}
\setvalue{site_biblestudytools_com_french}{0}
\setvalue{site_biblestudytools_com_italian}{0}
\setvalue{site_biblestudytools_com_polish}{0}
\setvalue{site_biblestudytools_com_spanish}{0}
\setvalue{site_biblestudytools_com_total}{37}
\setvalue{site_bibliepolskie_pl_english}{0}
\setvalue{site_bibliepolskie_pl_french}{0}
\setvalue{site_bibliepolskie_pl_italian}{0}
\setvalue{site_bibliepolskie_pl_polish}{20}
\setvalue{site_bibliepolskie_pl_spanish}{0}
\setvalue{site_bibliepolskie_pl_total}{20}
\setvalue{site_crossbible_com_english}{12}
\setvalue{site_crossbible_com_french}{1}
\setvalue{site_crossbible_com_italian}{0}
\setvalue{site_crossbible_com_polish}{0}
\setvalue{site_crossbible_com_spanish}{1}
\setvalue{site_crossbible_com_total}{14}
\setvalue{site_ebible_org_english}{24}
\setvalue{site_ebible_org_french}{4}
\setvalue{site_ebible_org_italian}{2}
\setvalue{site_ebible_org_polish}{2}
\setvalue{site_ebible_org_spanish}{5}
\setvalue{site_ebible_org_total}{37}
\setvalue{site_jw_org_english}{1}
\setvalue{site_jw_org_french}{1}
\setvalue{site_jw_org_italian}{1}
\setvalue{site_jw_org_polish}{1}
\setvalue{site_jw_org_spanish}{1}
\setvalue{site_jw_org_total}{5}
\setvalue{site_laparola_net_english}{9}
\setvalue{site_laparola_net_french}{3}
\setvalue{site_laparola_net_italian}{11}
\setvalue{site_laparola_net_polish}{0}
\setvalue{site_laparola_net_spanish}{2}
\setvalue{site_laparola_net_total}{25}
\setvalue{site_obohu_cz_english}{29}
\setvalue{site_obohu_cz_french}{17}
\setvalue{site_obohu_cz_italian}{3}
\setvalue{site_obohu_cz_polish}{9}
\setvalue{site_obohu_cz_spanish}{16}
\setvalue{site_obohu_cz_total}{74}
\setvalue{copyright_public_domain_english}{151}
\setvalue{copyright_open_license_english}{40}
\setvalue{copyright_copyrighted_english}{199}
\setvalue{copyright_public_domain_french}{42}
\setvalue{copyright_open_license_french}{2}
\setvalue{copyright_copyrighted_french}{34}
\setvalue{copyright_public_domain_italian}{13}
\setvalue{copyright_open_license_italian}{0}
\setvalue{copyright_copyrighted_italian}{20}
\setvalue{copyright_public_domain_polish}{13}
\setvalue{copyright_open_license_polish}{12}
\setvalue{copyright_copyrighted_polish}{23}
\setvalue{copyright_public_domain_spanish}{18}
\setvalue{copyright_open_license_spanish}{11}
\setvalue{copyright_copyrighted_spanish}{73}
\setvalue{copyright_public_domain_total}{237}
\setvalue{copyright_open_license_total}{65}
\setvalue{copyright_copyrighted_total}{349}
\setvalue{copyright_grand_total}{651}
\setvalue{booklength_n_comparisons}{10}
\setvalue{booklength_eng_fra_p}{0.002}
\setvalue{booklength_eng_fra_p_adj}{0.015}
\setvalue{booklength_eng_ita_p}{0.87}
\setvalue{booklength_eng_ita_p_adj}{1.00}
\setvalue{booklength_eng_pol_p}{0.045}
\setvalue{booklength_eng_pol_p_adj}{0.45}
\setvalue{booklength_eng_spa_p}{0.81}
\setvalue{booklength_eng_spa_p_adj}{1.00}
\setvalue{booklength_fra_ita_p}{0.030}
\setvalue{booklength_fra_ita_p_adj}{0.30}
\setvalue{booklength_fra_pol_p}{0.001}
\setvalue{booklength_fra_pol_p_adj}{0.012}
\setvalue{booklength_fra_spa_p}{0.004}
\setvalue{booklength_fra_spa_p_adj}{0.038}
\setvalue{booklength_ita_pol_p}{0.20}
\setvalue{booklength_ita_pol_p_adj}{1.00}
\setvalue{booklength_ita_spa_p}{0.96}
\setvalue{booklength_ita_spa_p_adj}{1.00}
\setvalue{booklength_pol_spa_p}{0.10}
\setvalue{booklength_pol_spa_p_adj}{1.00}
\setvalue{booklength_mean_n_eng}{194}
\setvalue{booklength_mean_n_fra}{41}
\setvalue{booklength_mean_n_ita}{17}
\setvalue{booklength_mean_n_pol}{29}
\setvalue{booklength_mean_n_spa}{53}
\setvalue{booklength_mean_eng_fra_p}{0.000}
\setvalue{booklength_mean_eng_fra_p_adj}{0.000}
\setvalue{booklength_mean_eng_ita_p}{0.38}
\setvalue{booklength_mean_eng_ita_p_adj}{1.00}
\setvalue{booklength_mean_eng_pol_p}{0.000}
\setvalue{booklength_mean_eng_pol_p_adj}{0.000}
\setvalue{booklength_mean_eng_spa_p}{0.11}
\setvalue{booklength_mean_eng_spa_p_adj}{1.00}
\setvalue{booklength_mean_fra_ita_p}{0.000}
\setvalue{booklength_mean_fra_ita_p_adj}{0.000}
\setvalue{booklength_mean_fra_pol_p}{0.000}
\setvalue{booklength_mean_fra_pol_p_adj}{0.000}
\setvalue{booklength_mean_fra_spa_p}{0.000}
\setvalue{booklength_mean_fra_spa_p_adj}{0.000}
\setvalue{booklength_mean_ita_pol_p}{0.001}
\setvalue{booklength_mean_ita_pol_p_adj}{0.012}
\setvalue{booklength_mean_ita_spa_p}{0.77}
\setvalue{booklength_mean_ita_spa_p_adj}{1.00}
\setvalue{booklength_mean_pol_spa_p}{0.000}
\setvalue{booklength_mean_pol_spa_p_adj}{0.001}
\setvalue{booklength_mean_n_sig}{7}
\setvalue{sim_all_n}{85657}
\setvalue{sim_all_p10}{0.388}
\setvalue{sim_all_p50}{0.531}
\setvalue{sim_all_p90}{0.712}
\setvalue{sim_cid_n}{2173}
\setvalue{sim_cid_p10}{0.910}
\setvalue{sim_cid_p50}{0.990}
\setvalue{sim_cid_p90}{1.000}
\setvalue{sim_ver_n}{803}
\setvalue{sim_ver_p10}{0.967}
\setvalue{sim_ver_p50}{0.996}
\setvalue{sim_ver_p90}{1.000}
\setvalue{sim_diff_cid_n}{83484}
\setvalue{sim_diff_cid_p10}{0.387}
\setvalue{sim_diff_cid_p50}{0.527}
\setvalue{sim_diff_cid_p90}{0.700}
\setvalue{sim_english_cid_n}{1673}
\setvalue{sim_english_cid_p10}{0.944}
\setvalue{sim_english_cid_p50}{0.991}
\setvalue{sim_english_cid_p90}{1.000}
\setvalue{sim_english_ver_n}{556}
\setvalue{sim_english_ver_p10}{0.963}
\setvalue{sim_english_ver_p50}{0.995}
\setvalue{sim_english_ver_p90}{1.000}
\setvalue{sim_french_cid_n}{119}
\setvalue{sim_french_cid_p10}{0.903}
\setvalue{sim_french_cid_p50}{0.986}
\setvalue{sim_french_cid_p90}{1.000}
\setvalue{sim_french_ver_n}{91}
\setvalue{sim_french_ver_p10}{0.963}
\setvalue{sim_french_ver_p50}{0.988}
\setvalue{sim_french_ver_p90}{1.000}
\setvalue{sim_italian_cid_n}{56}
\setvalue{sim_italian_cid_p10}{0.835}
\setvalue{sim_italian_cid_p50}{0.991}
\setvalue{sim_italian_cid_p90}{1.000}
\setvalue{sim_italian_ver_n}{26}
\setvalue{sim_italian_ver_p10}{0.835}
\setvalue{sim_italian_ver_p50}{0.998}
\setvalue{sim_italian_ver_p90}{1.000}
\setvalue{sim_polish_cid_n}{47}
\setvalue{sim_polish_cid_p10}{0.986}
\setvalue{sim_polish_cid_p50}{0.998}
\setvalue{sim_polish_cid_p90}{1.000}
\setvalue{sim_polish_ver_n}{38}
\setvalue{sim_polish_ver_p10}{0.993}
\setvalue{sim_polish_ver_p50}{0.999}
\setvalue{sim_polish_ver_p90}{1.000}
\setvalue{sim_spanish_cid_n}{278}
\setvalue{sim_spanish_cid_p10}{0.755}
\setvalue{sim_spanish_cid_p50}{0.956}
\setvalue{sim_spanish_cid_p90}{1.000}
\setvalue{sim_spanish_ver_n}{92}
\setvalue{sim_spanish_ver_p10}{0.990}
\setvalue{sim_spanish_ver_p50}{0.999}
\setvalue{sim_spanish_ver_p90}{1.000}
\setvalue{sim_diachronic_english_lex_n}{18528}
\setvalue{sim_diachronic_english_lex_r2}{0.0197}
\setvalue{sim_diachronic_english_lex_midyear_beta}{-0.03402}
\setvalue{sim_diachronic_english_lex_midyear_p}{0.000}
\setvalue{sim_diachronic_english_lex_midyear_p_fmt}{< .001}
\setvalue{sim_diachronic_english_lex_midyear_sig}{yes}
\setvalue{sim_diachronic_french_lex_n}{814}
\setvalue{sim_diachronic_french_lex_r2}{0.1341}
\setvalue{sim_diachronic_french_lex_midyear_beta}{-0.05848}
\setvalue{sim_diachronic_french_lex_midyear_p}{0.000}
\setvalue{sim_diachronic_french_lex_midyear_p_fmt}{< .001}
\setvalue{sim_diachronic_french_lex_midyear_sig}{yes}
\setvalue{sim_diachronic_italian_lex_n}{129}
\setvalue{sim_diachronic_italian_lex_r2}{0.0585}
\setvalue{sim_diachronic_italian_lex_midyear_beta}{-0.05390}
\setvalue{sim_diachronic_italian_lex_midyear_p}{0.011}
\setvalue{sim_diachronic_italian_lex_midyear_p_fmt}{= 0.011}
\setvalue{sim_diachronic_italian_lex_midyear_sig}{yes}
\setvalue{sim_diachronic_polish_lex_n}{402}
\setvalue{sim_diachronic_polish_lex_r2}{0.2367}
\setvalue{sim_diachronic_polish_lex_midyear_beta}{-0.07290}
\setvalue{sim_diachronic_polish_lex_midyear_p}{0.000}
\setvalue{sim_diachronic_polish_lex_midyear_p_fmt}{< .001}
\setvalue{sim_diachronic_polish_lex_midyear_sig}{yes}
\setvalue{sim_diachronic_spanish_lex_n}{1348}
\setvalue{sim_diachronic_spanish_lex_r2}{0.1169}
\setvalue{sim_diachronic_spanish_lex_midyear_beta}{-0.10069}
\setvalue{sim_diachronic_spanish_lex_midyear_p}{0.000}
\setvalue{sim_diachronic_spanish_lex_midyear_p_fmt}{< .001}
\setvalue{sim_diachronic_spanish_lex_midyear_sig}{yes}
\setvalue{sim_diachronic_english_sem_n}{18528}
\setvalue{sim_diachronic_english_sem_r2}{0.0537}
\setvalue{sim_diachronic_english_sem_midyear_beta}{-0.00340}
\setvalue{sim_diachronic_english_sem_midyear_p}{0.000}
\setvalue{sim_diachronic_english_sem_midyear_p_fmt}{< .001}
\setvalue{sim_diachronic_english_sem_midyear_sig}{yes}
\setvalue{sim_diachronic_french_sem_n}{814}
\setvalue{sim_diachronic_french_sem_r2}{0.0297}
\setvalue{sim_diachronic_french_sem_midyear_beta}{-0.00431}
\setvalue{sim_diachronic_french_sem_midyear_p}{0.000}
\setvalue{sim_diachronic_french_sem_midyear_p_fmt}{< .001}
\setvalue{sim_diachronic_french_sem_midyear_sig}{yes}
\setvalue{sim_diachronic_italian_sem_n}{129}
\setvalue{sim_diachronic_italian_sem_r2}{0.1198}
\setvalue{sim_diachronic_italian_sem_midyear_beta}{-0.00987}
\setvalue{sim_diachronic_italian_sem_midyear_p}{0.000}
\setvalue{sim_diachronic_italian_sem_midyear_p_fmt}{< .001}
\setvalue{sim_diachronic_italian_sem_midyear_sig}{yes}
\setvalue{sim_diachronic_polish_sem_n}{402}
\setvalue{sim_diachronic_polish_sem_r2}{0.2790}
\setvalue{sim_diachronic_polish_sem_midyear_beta}{-0.00390}
\setvalue{sim_diachronic_polish_sem_midyear_p}{0.001}
\setvalue{sim_diachronic_polish_sem_midyear_p_fmt}{= 0.001}
\setvalue{sim_diachronic_polish_sem_midyear_sig}{yes}
\setvalue{sim_diachronic_spanish_sem_n}{1348}
\setvalue{sim_diachronic_spanish_sem_r2}{0.0915}
\setvalue{sim_diachronic_spanish_sem_midyear_beta}{-0.00805}
\setvalue{sim_diachronic_spanish_sem_midyear_p}{0.000}
\setvalue{sim_diachronic_spanish_sem_midyear_p_fmt}{< .001}
\setvalue{sim_diachronic_spanish_sem_midyear_sig}{yes}
\setvalue{sim_diachronic_lex_n_sig_neg}{5}
\setvalue{sim_diachronic_lex_all_p_below_001}{no}
\setvalue{sim_diachronic_lex_min_beta_lang}{english}
\setvalue{sim_diachronic_lex_max_beta_lang}{spanish}
\setvalue{sim_diachronic_lex_r2_min}{0.02}
\setvalue{sim_diachronic_lex_r2_max}{0.24}
\setvalue{sim_diachronic_sem_n_sig_neg}{5}
\setvalue{sim_diachronic_sem_all_p_below_001}{no}
\setvalue{sim_diachronic_sem_min_beta_lang}{english}
\setvalue{sim_diachronic_sem_max_beta_lang}{italian}
\setvalue{sim_diachronic_sem_r2_min}{0.03}
\setvalue{sim_diachronic_sem_r2_max}{0.28}
\setvalue{sim_diachronic_r2_min}{0.02}
\setvalue{sim_diachronic_r2_max}{0.28}
\setvalue{xsim_english_french_mean}{0.862}
\setvalue{xsim_english_french_n}{30420}
\setvalue{xsim_english_italian_mean}{0.842}
\setvalue{xsim_english_italian_n}{12870}
\setvalue{xsim_english_polish_mean}{0.790}
\setvalue{xsim_english_polish_n}{18720}
\setvalue{xsim_english_spanish_mean}{0.854}
\setvalue{xsim_english_spanish_n}{39780}
\setvalue{xsim_french_italian_mean}{0.841}
\setvalue{xsim_french_italian_n}{2574}
\setvalue{xsim_french_polish_mean}{0.766}
\setvalue{xsim_french_polish_n}{3744}
\setvalue{xsim_french_spanish_mean}{0.827}
\setvalue{xsim_french_spanish_n}{7956}
\setvalue{xsim_italian_polish_mean}{0.776}
\setvalue{xsim_italian_polish_n}{1584}
\setvalue{xsim_italian_spanish_mean}{0.863}
\setvalue{xsim_italian_spanish_n}{3366}
\setvalue{xsim_polish_spanish_mean}{0.803}
\setvalue{xsim_polish_spanish_n}{4896}
\setvalue{xsim_english_within_mean}{0.943}
\setvalue{xsim_french_within_mean}{0.965}
\setvalue{xsim_italian_within_mean}{0.965}
\setvalue{xsim_polish_within_mean}{0.955}
\setvalue{xsim_spanish_within_mean}{0.947}
\setvalue{xsim_within_mean}{0.955}
\setvalue{xsim_within_min}{0.943}
\setvalue{xsim_within_max}{0.965}
\setvalue{xsim_within_min_lang}{english}
\setvalue{xsim_within_max_lang}{french}
\setvalue{xsim_cross_mean}{0.823}
\setvalue{xsim_cross_min}{0.766}
\setvalue{xsim_cross_max}{0.863}
\setvalue{xsim_cross_min_pair}{french--polish}
\setvalue{xsim_cross_max_pair}{italian--spanish}

\usepackage{iftex}

\iftutex
  \usepackage{fontspec}
\else
  \usepackage[NHE8,LGR,T1]{fontenc}
\fi

\ifluatex
  \usepackage[bidi=basic, provide*=*]{babel}
\else
  \usepackage[bidi=default, provide*=*]{babel}
\fi

\babelprovide[import, main]{american}
\babelprovide[import, onchar=ids fonts]{polytonicgreek}
\babelprovide[import, onchar=ids fonts]{hebrew}

\title{Targum --- A Multilingual New Testament Translation Corpus}

\name{Maciej Rapacz, Aleksander Smywiński-Pohl}

\address{AGH University of Kraków \\
         \{mrapacz, apohllo\}@agh.edu.pl\\}

\begin{document}

\abstract{
    Many European languages possess rich biblical translation histories, yet existing corpora --- in prioritizing linguistic breadth --- often fail to capture this depth. To address this gap, we introduce a multilingual corpus of \data{total_translations} New Testament translations, of which \data{unique_translations} are unique, spanning five languages with 2.4--5.0$\times$ more translations per language than any prior corpus: English (\data{unique_english} unique versions from \data{total_english} total), French (\data{unique_french} from \data{total_french}), Italian (\data{unique_italian} from \data{total_italian}), Polish (\data{unique_polish} from \data{total_polish}), and Spanish (\data{unique_spanish} from \data{total_spanish}). Aggregated from \data{num_online_sources} online biblical libraries and one preexisting corpus, each translation is annotated with metadata that maps the text to a standardized identifier for the work, its specific edition, and its year of revision. This canonicalization allows researchers to define "uniqueness" for their own needs: they can perform micro-level analyses on translation families, such as the KJV lineage, or conduct macro-level studies by deduplicating closely related texts. By providing the first multilingual resource with sufficient depth per language for flexible, multilevel analysis, the corpus fills a gap in the quantitative study of translation history. \\ \newline  }

\maketitleabstract

\section{Introduction}

In the data-scarce era of early Natural Language Processing, the Bible was a widely used parallel corpus. Its utility for methods such as statistical machine translation (SMT) was based on the assumption that the verses between translations were semantically equivalent \cite{mueller-etal-2020-analysis}. The current landscape of NLP, characterized by terabyte-scale datasets and large language models, redefines this utility. The Bible's volume is negligible for training modern foundation models, and its specific thematic and stylistic registers are not representative of general language use. This shift reframes the main application of the corpus: from being data \textit{for} AI systems to being an object of study \textit{by} AI systems, enabling large-scale quantitative analysis of textual choices that reflect its cultural and theological history. Far from a relic of the past, Bible translation remains an active enterprise: a substantial share of the translations in our corpus date from the 21st century, underscoring the continued demand for new renderings across confessional and stylistic lines.

We introduce \texttt{Targum},\footnote{\url{https://github.com/mrapacz/targum-corpus}} a corpus designed to prioritize depth over linguistic breadth. \texttt{Targum} provides coverage for five languages with extensive translation histories --- English, French, Italian, Polish, and Spanish. This depth supports computational analysis of a wide spectrum of historical periods and confessional traditions, treating translation variants that heuristic methods previously discarded as a primary object of study.

The corpus is named after the Targum (\foreignlanguage{hebrew}{תרגום}, ``translation''), the tradition of ancient Aramaic Bible translations that combined literal rendering with dynamic expansion to serve their audience \cite{alexander1992targum, schuhlein1912targum_online}. Our corpus is designed to span this same spectrum --- from the most formal, literal renderings to the most dynamic paraphrases.

The main contributions of this paper are two-fold. First, we introduce \texttt{Targum}, a corpus of \data{total_translations} New Testament translations (\data{unique_translations} unique) across five European languages, providing 2.4--5.0$\times$ the depth of any prior resource per language. Second, we verify and normalize the metadata for each translation, including its publication history and edition identity. We first position our work with respect to existing corpora, then detail the methodology for corpus construction and metadata normalization, describe the resulting corpus and its distribution format, present a statistical analysis of the resource, and conclude by discussing new research directions.

\section{Related Work}

\begin{table*}[htbp]
    \centering
    \begin{tabular}{lrrrrrr}
            \toprule
            Language & English & French & Italian & Polish & Spanish & \textbf{Total} \\
            \midrule
            Corpus &  &  &  &  &  &  \\
            \citet{christodouloupoulos2015massively} & \data{prior_christodouloupoulos_english} & \data{prior_christodouloupoulos_french} & \data{prior_christodouloupoulos_italian} & \data{prior_christodouloupoulos_polish} & \data{prior_christodouloupoulos_spanish} & \textbf{\data{prior_christodouloupoulos_total}} \\
            \citet{akerman2023ebible} & \data{prior_ebible_english} & \data{prior_ebible_french} & \data{prior_ebible_italian} & \data{prior_ebible_polish} & \data{prior_ebible_spanish} & \textbf{\data{prior_ebible_total}} \\
            \citet{mayer_creating_2014} & \data{prior_mayer_english} & \data{prior_mayer_french} & \data{prior_mayer_italian} & \data{prior_mayer_polish} & \data{prior_mayer_spanish} & \textbf{\data{prior_mayer_total}} \\
            \midrule
            Targum & \data{total_english} & \data{total_french} & \data{total_italian} & \data{total_polish} & \data{total_spanish} & \textbf{\data{total_translations}} \\
            Targum (no duplicates) & \data{unique_english} & \data{unique_french} & \data{unique_italian} & \data{unique_polish} & \data{unique_spanish} & \textbf{\data{unique_translations}} \\
            \bottomrule
        \end{tabular}
    \caption{The number of translations for each language in the Bible translation corpora.}
    \label{tab:corpus_counts}
\end{table*}

The Bible is a frequently used text for creating multilingual corpora due to its verse structure, which provides de facto parallel sentence-level alignment in a large number of languages \cite{mueller-etal-2020-analysis}. Research using these corpora has followed two main paradigms. The most common has been \textit{instrumental}, using the text as a convenient dataset to bootstrap NLP tools for low-resource languages, such as training machine translation systems \cite{guzman-etal-2019-flores} or creating part-of-speech taggers via annotation projection \cite{nicolai-yarowsky-2019-learning, nicolai-etal-2020-fine}. In contrast, a more recent paradigm, often aligned with the digital humanities, treats the biblical text as the \textit{primary object of study}. This line of research applies computational methods to perform a diachronic analysis of language change or to study stylistic variation between translations. Our work is situated within this second paradigm.

A significant body of work has focused on creating biblical corpora that maximize linguistic breadth. Over time, a series of progressively larger collections were introduced, starting with foundational resources such as those from \citet{christodouloupoulos2015massively} and \citet{mayer_creating_2014}. This trend culminated in the Johns Hopkins University Bible Corpus, which expanded coverage to over 1,600 languages to support tasks such as typological analysis and low-resource NLP \cite{mccarthy2020johns}. The Parallel Bible Corpus \cite{mayer_creating_2014} has also served as a foundation for typological studies, such as the computational investigation of tense marking across over 1,000 languages \cite{asgari-schutze-2017-past}. The state of these digital resources, however, is often dynamic; for instance, while some corpora like that of \citet{mayer_creating_2014} are actively maintained and have grown since their initial publication, the Johns Hopkins corpus is no longer publicly available. A more recent effort, the eBible Corpus, provides another standardized and accessible resource \cite{akerman2023ebible}. While these collections offer wide breadth, their depth for any single language remains limited. Table \ref{tab:corpus_counts} quantifies this gap: against the largest prior resource for each language \cite{mayer_creating_2014}, \texttt{Targum} contains \data{unique_english} unique English versions (\data{multiplier_english}$\times$ increase), \data{unique_french} French (\data{multiplier_french}$\times$), \data{unique_italian} Italian (\data{multiplier_italian}$\times$), \data{unique_polish} Polish (\data{multiplier_polish}$\times$), and \data{unique_spanish} Spanish (\data{multiplier_spanish}$\times$).

Although most work has focused on breadth, some projects have curated collections with depth for specific scholarly questions. A prime example is the EDGeS Diachronic Bible Corpus, a resource containing 36 translations across four Germanic languages designed to enable the ``longitudinal and contrastive study of complex verb constructions'' \cite{bouma2020edges}. Similarly, \citet{carlson2018evaluating} compiles 34 English versions for the specific task of evaluating the transfer of prose style. Our work extends this depth-focused approach to a multilingual setting. Although these prior corpora were built for predefined research questions, \texttt{Targum} is designed as a \textit{general-purpose resource for comparative analysis}, letting researchers construct subcorpora tailored to specific inquiries across multiple languages and traditions.

\section{Methodology}

\subsection{Data Acquisition and Processing}

\begin{table*}[htbp]
    \centering
    \begin{tabular}{lrrrrrr}
    \toprule
    Language & English & French & Italian & Polish & Spanish & \textbf{Total} \\
    Source &  &  &  &  &  &  \\
    \midrule
    bible.audio & 3 & 15 & 0 & 0 & 1 & \textbf{19} \\
    bible.com & 75 & 19 & 7 & 8 & 28 & \textbf{137} \\
    bible.is & 18 & 8 & 1 & 2 & 15 & \textbf{44} \\
    biblegateway.com & 64 & 4 & 5 & 3 & 19 & \textbf{95} \\
    biblehub.com & 46 & 0 & 0 & 0 & 0 & \textbf{46} \\
    bibles.org & 72 & 6 & 3 & 3 & 14 & \textbf{98} \\
    biblestudytools.com & 37 & 0 & 0 & 0 & 0 & \textbf{37} \\
    bibliepolskie.pl & 0 & 0 & 0 & 20 & 0 & \textbf{20} \\
    crossbible.com & 12 & 1 & 0 & 0 & 1 & \textbf{14} \\
    ebible.org & 24 & 4 & 2 & 2 & 5 & \textbf{37} \\
    jw.org & 1 & 1 & 1 & 1 & 1 & \textbf{5} \\
    laparola.net & 9 & 3 & 11 & 0 & 2 & \textbf{25} \\
    obohu.cz & 29 & 17 & 3 & 9 & 16 & \textbf{74} \\
    \midrule
    \textbf{Total} & \textbf{390} & \textbf{78} & \textbf{33} & \textbf{48} & \textbf{102} & \textbf{651} \\
    \bottomrule
\end{tabular}
    \caption{Number of translations sourced from each site for each language in the \texttt{Targum} corpus.}
    \label{tab:source_counts}
\end{table*}

The construction of the corpus began with the aggregation of translations from a diverse set of sources, including \data{num_online_sources} online biblical libraries and one preexisting digital corpus. The latter comprises translations published by eBible.org \cite{akerman2023ebible}, which we sourced from its public GitHub repository; in our source table (Table~\ref{tab:source_counts}) these appear under the \textit{ebible.org} site. We identified well-known, large-scale aggregators (e.g., Bible Gateway, Bible Hub) to ensure broad coverage, and supplemented these with language-specific repositories (e.g., \texttt{bibliepolskie.pl} for Polish, \texttt{laparola.net} for Italian) to achieve deep coverage in our target languages. The acquisition pipeline consists of three stages.

First, an \textit{indexer} inventories all available translations on each source website, capturing metadata such as source-specific identifiers and translation names, producing a stable, timestamped snapshot of each site's collection.

Second, a \textit{scraper} uses this index to download the full text of each translation, caching the raw source files (e.g., HTML or JSON) for each of the New Testament's 260 chapters without modification. Separating download from parsing allows iterative refinement of extraction logic offline, prevents repeated requests to source websites, and guarantees reproducibility.

Third, a \textit{parser} processes the cached files to extract verse-level text. A key challenge at this stage is the structural variation in verse notation. While many translations align to a simple one-to-one mapping, more dynamic versions often introduce non-standard segments (e.g., 6a, 6b) or consolidated ranges (e.g., 6--8). Our parser handles these cases. The final output is stored as a newline-delimited JSON file (JSONL) that preserves the detailed verse segmentation; the structure of these files is documented in Appendix~\ref{sec:appendix_hierarchy}.

\subsection{Canonicalization and Metadata Normalization}
The scraped data is inconsistent across sources; the same translation often appears with different names and inconsistent metadata (e.g., publication year). We resolved this by mapping each scraped instance to a grouping identifier that unifies all instances of the same translation work. This metadata is stored in YAML files, where each translation is assigned a \texttt{canonical\_id} and a \texttt{canonical\_revision\_year}. The \texttt{canonical\_id} groups together all instances that represent the same translation work regardless of their source, but it does not model the full complexity of translation lineage. For example, all scraped copies of the King James Version share one \texttt{canonical\_id}, while the KJV's derivative translations (e.g., NKJV, RV, ASV) receive separate identifiers. We also record a \texttt{canonical\_version} string that identifies the specific edition: while it often reduces to the year (e.g., \texttt{"1769"}), it sometimes encodes richer distinctions such as \texttt{"2nd-edition"}, \texttt{"1989-anglicised-catholic"}, or \texttt{"1989-americanized"}. Two instances sharing the same \texttt{canonical\_id} and \texttt{canonical\_version} from different sources should be near-identical texts (the same edition scraped from multiple sites). Within each group, the \texttt{canonical\_revision\_year} is an integer that places the edition on the time axis. For both the copyright status and version fields we store accompanying reasoning fields that document the evidence behind each decision. We acknowledge that some translation histories are highly complex, involving works that draw from multiple predecessors; we plan to address these with more granular metadata in future work.

Determining these values required extensive external research into publication histories, prefaces, and publisher catalogs. While we initially explored AI assistants for automated metadata extraction, they proved unreliable under manual review, often failing to distinguish between closely-related revisions or misidentifying publication details. We therefore performed the normalization manually. We note that this task is factual cataloguing --- mapping a scraped text to its real-world publication identity --- rather than a subjective linguistic judgment. Each case is resolved by tracing documentary evidence (title pages, publisher records, preface dates), and the reasoning is recorded alongside each annotation to support future reassessment. The cross-source validation described in the following section provides an additional safeguard: translations grouped under the same identifier are checked for textual consistency using similarity metrics.

\begin{table*}[htbp]
    \centering
    \begin{tabular}{lrrrrrr}
    \toprule
    Language & English & French & Italian & Polish & Spanish & \textbf{Total} \\
    Copyright Status &  &  &  &  &  &  \\
    \midrule
    Public Domain & 151 & 42 & 13 & 13 & 18 & \textbf{237} \\
    Open License & 40 & 2 & 0 & 12 & 11 & \textbf{65} \\
    Copyrighted & 199 & 34 & 20 & 23 & 73 & \textbf{349} \\
    \midrule
    \textbf{Total} & \textbf{390} & \textbf{78} & \textbf{33} & \textbf{48} & \textbf{102} & \textbf{651} \\
    \bottomrule
\end{tabular}
    \caption{Copyright status distribution of translations in the \texttt{Targum} corpus by language.}
    \label{tab:copyright_stats}
\end{table*}

\subsection{Quality Control via Cross-Source Validation}
The aggregation process naturally yields multiple copies of the same canonical text, particularly for widely-distributed versions. Instead of viewing these as duplicates to be discarded, we treat their presence as a central feature of our quality control methodology.

Once scraped translations are assigned the same \texttt{canonical\_id} and \texttt{canonical\_version}, we conduct a systematic, side-by-side comparison of their texts using lexical similarity metrics. Such metrics are sensitive to the single-character variations that distinguish editions or reflect transcription errors. This comparison reliably spots discrepancies, which expose one of two issues: a subtle error in our parsing logic, or a pre-existing data error on the source website, such as a typo or a missing verse. To resolve these differences and identify the correct edition, we employ several verification techniques, including comparing the text against scanned historical documents and using search engines to find quotations of a specific divergent phrase, which can often be traced back to a particular publication. This iterative process allows us to correct our parsers and improve the textual accuracy of the entire collection.

\section{The Targum Corpus}\label{sec:corpus_description}

\subsection{Sources and Scale}

Table \ref{tab:source_counts} details the provenance of the \data{total_translations} translations collected by the pipeline described above. Coverage is distributed across both large-scale international aggregators and specialized, language-specific repositories. English is well-represented across numerous sites, with \textit{bible.com}, \textit{biblegateway.com}, and \textit{bibles.org} as the largest sources. \textit{bible.com} also serves as the single largest source for Spanish translations. For other languages, a combination of sources proved most effective; French, for instance, drew heavily from \textit{bible.com}, \textit{bible.audio}, and \textit{obohu.cz}. In contrast, the collections for Polish and Italian are led by their respective national sites, \textit{bibliepolskie.pl} and \textit{laparola.net}. Achieving broad coverage required combining international aggregators with language-specific repositories.

\subsection{Corpus Structure and Distribution Format}\label{sec:corpus_structure}

The corpus is not a single, pre-filtered dataset where duplicates have been removed. Instead, we deliver the complete archive of all collected translations, accompanied by a metadata index recording the provenance and specific edition of each text.

This structure gives researchers full control rather than imposing a pre-defined notion of what constitutes a ``duplicate.'' The rich metadata allows users to define uniqueness on their own terms, tailored to their specific research question. For example, a scholar studying Polish translations might prioritize copies from the language-specific \textit{bibliepolskie.pl} source. A researcher analyzing the evolution of the KJV family would treat each revision as a distinct entity. Conversely, a user performing a broad, cross-lingual study might deduplicate aggressively, keeping only one instance per \texttt{canonical\_id}.

Each translation instance is described by: source provenance (\texttt{site}), language (\texttt{iso}), source-specific identifiers (\texttt{translation\_id}, \texttt{translation\_name}, \texttt{translation\_abbr}), canonical identity (\texttt{canonical\_id}, \texttt{canonical\_version}, \texttt{canonical\_year}), completeness statistics (\texttt{num\_chapters}, \texttt{num\_verses}, \texttt{num\_words}), and \texttt{copyright\_status} with documented reasoning. A complete field reference with descriptions and examples is provided in Appendix~\ref{sec:appendix_metadata}; the full file hierarchy is documented in Appendix~\ref{sec:appendix_hierarchy}.

In addition to the text files and metadata index, the corpus distributes several pre-computed resources: a \textit{book coverage} file recording which of the 27 New Testament books are present in each translation; chapter- and verse-level vector embeddings using the Qwen3-Embedding model \cite{qwen3embedding} in two sizes (0.6B and 8B), selected for its multilingual performance and long-context capabilities; and pre-computed pairwise intra-lingual similarity scores for all translation pairs within each language, covering both semantic similarity (cosine distance on the 8B embeddings) and lexical similarity (Levenshtein), computed at the chapter level. These pre-computed resources support immediate downstream analysis without expensive recomputation, and allow research on copyrighted translations without distributing the source texts. Table \ref{tab:copyright_stats} presents the copyright status distribution of translations across languages.

\begin{figure*}[htbp]
    \centering
    \includegraphics[width=\linewidth]{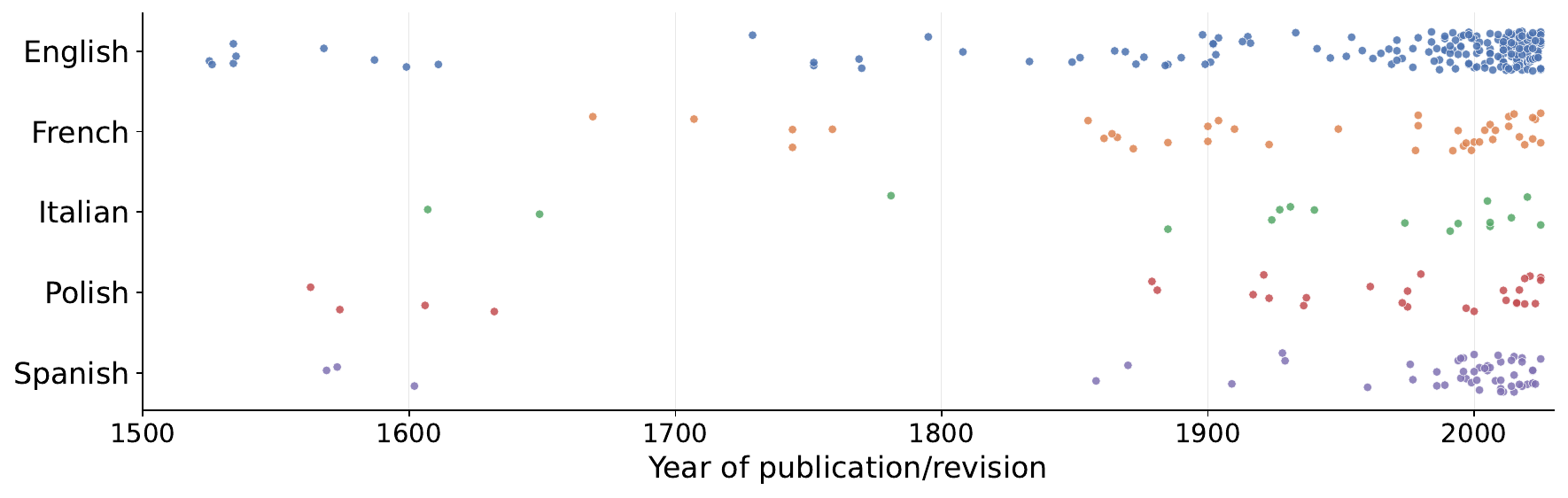}
    \caption{Diachronic distribution of translations per language. Each dot represents a distinct (\texttt{canonical\_id}, \texttt{canonical\_version}) pair.}
    \label{fig:diachronic_dist}
\end{figure*}

\section{Corpus Analysis}\label{sec:corpus_analysis}

\subsection{Diachronic Distribution}
The depth of the corpus is not only numerical but also historical. Figure \ref{fig:diachronic_dist} illustrates the temporal distribution of the unique translations based on their revision year. Across all five languages, the distribution is skewed to the right, with a significant concentration of translations published in the last 50 years. The historical scope of the corpus begins in the 16th-century Reformation and Counter-Reformation period, with English showing the most continuous and extensive history from that point forward. The plot also reveals notable historical gaps. For Polish and Spanish, there is a long period of silence between a small cluster of early 17th-century translations and a renewed burst of activity in the second half of the 19th century. Similarly, while our Italian collection includes a few early 17th-century texts, this period remains scarce, and the earliest French translation in the corpus is from the latter half of the 17th century.

\subsection{Book Length Distribution}
\begin{figure*}[htbp]
    \centering
    \includegraphics[width=\textwidth]{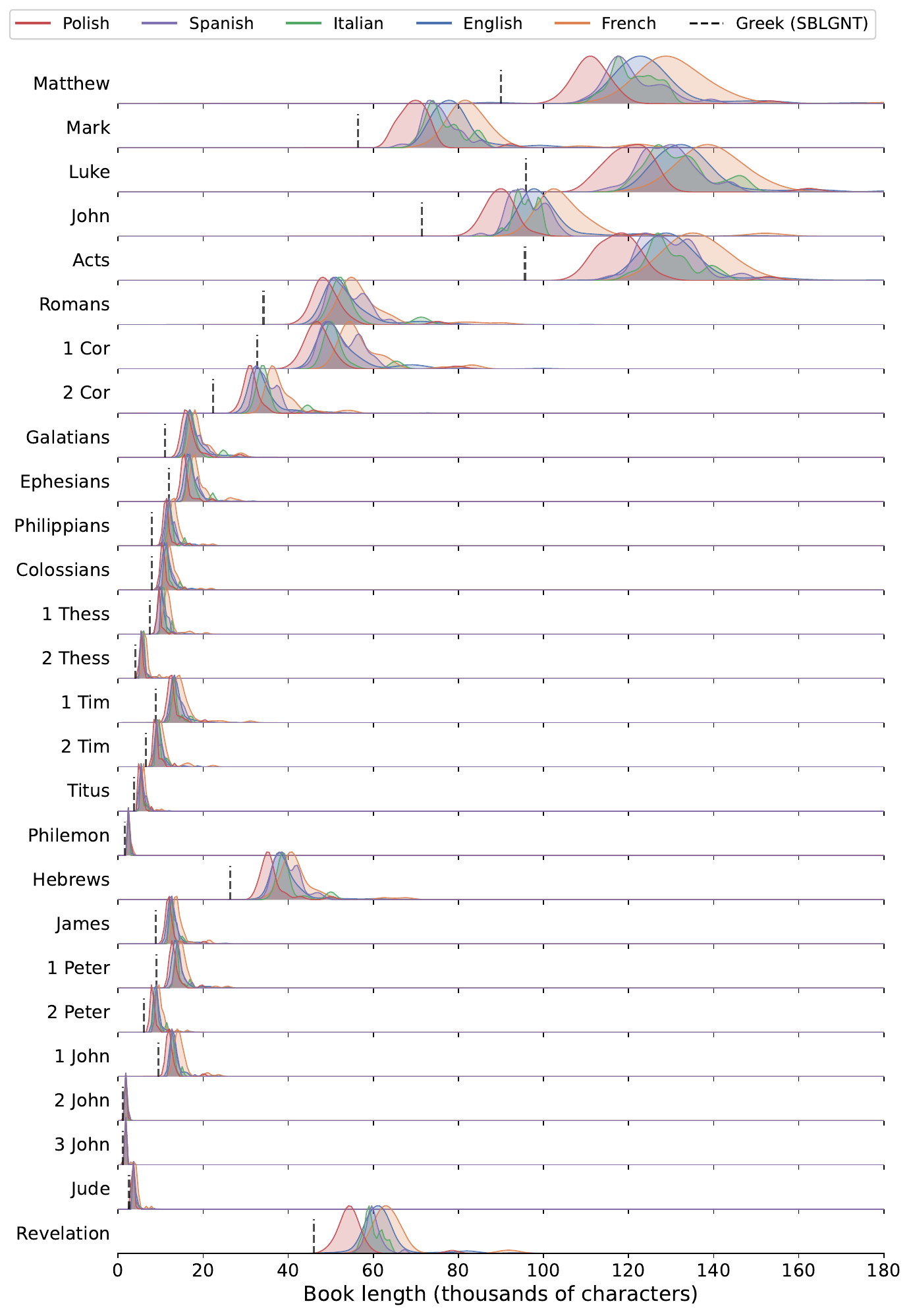}
    \caption{Distribution of book lengths (in characters) across the \data{unique_translations} unique translations for each of the 27 New Testament books, broken down by language. Each curve shows the kernel density estimate for one language. The dashed vertical line marks the length of the Greek source text (SBLGNT).}
    \label{fig:book_length_dist}
\end{figure*}

To characterize the structural properties of the corpus, we examined the distribution of book lengths (in characters) across the \data{unique_translations} unique translations (deduplicated by edition). For each of the 27 New Testament books, we aggregated the chapter-level text for every unique translation and computed the total character count. Figure \ref{fig:book_length_dist} presents these distributions as a ridgeline plot, with one density curve per language overlaid for each book. Matthew, Luke, John, and Acts are the longest books across all languages, while the shorter Epistles (e.g., Philemon, 2 John, 3 John, Jude) show tightly concentrated distributions. Notably, for every book the distributions of all five target languages are shifted to the right of the Greek source text (SBLGNT), indicated by the dashed vertical line, reflecting the expansion that naturally occurs in translation from a highly synthetic source language. While the distributions appear broadly similar and overlap substantially across languages, visual inspection suggests that French translations tend to be somewhat longer and Polish translations somewhat shorter than the other groups. To test this, we performed pairwise Mann-Whitney U tests over all book$\times$translation character counts with Bonferroni correction ($n=\data{booklength_n_comparisons}$ comparisons). The analysis confirms that French is significantly longer than English ($p_\text{adj}=\data{booklength_eng_fra_p_adj}$), Polish ($p_\text{adj}=\data{booklength_fra_pol_p_adj}$), and Spanish ($p_\text{adj}=\data{booklength_fra_spa_p_adj}$). The French--Italian difference does not survive correction ($p=\data{booklength_fra_ita_p}$, $p_\text{adj}=\data{booklength_fra_ita_p_adj}$). All other pairwise comparisons are non-significant after correction. Because the pooled test treats every book$\times$translation observation as independent, we also performed a secondary analysis on per-translation mean character counts (one observation per translation, $N$ ranging from \data{booklength_mean_n_ita} to \data{booklength_mean_n_eng}). This test, which removes within-translation book-to-book variance, yields \data{booklength_mean_n_sig} significant pairs: French is significantly longer than all four other languages (including Italian, $p_\text{adj}=\data{booklength_mean_fra_ita_p_adj}$), and Polish is significantly shorter than English ($p_\text{adj}=\data{booklength_mean_eng_pol_p_adj}$), French, Italian ($p_\text{adj}=\data{booklength_mean_ita_pol_p_adj}$), and Spanish ($p_\text{adj}=\data{booklength_mean_pol_spa_p_adj}$). English, Italian, and Spanish do not differ significantly from each other under either test. These differences likely reflect morphological and orthographic properties of each language (e.g., French tends to use more articles, prepositions, and accented characters; Polish uses fewer function words due to its rich case system) rather than translation philosophy.

\subsection{Pairwise Similarity Structure}
Given that the corpus contains numerous translations for each language, including multiple copies of widely-distributed versions, we quantify their similarity across all New Testament books using two comparison methods: semantic and lexical.

\begin{figure*}[htbp]
    \centering
    \includegraphics[width=\textwidth]{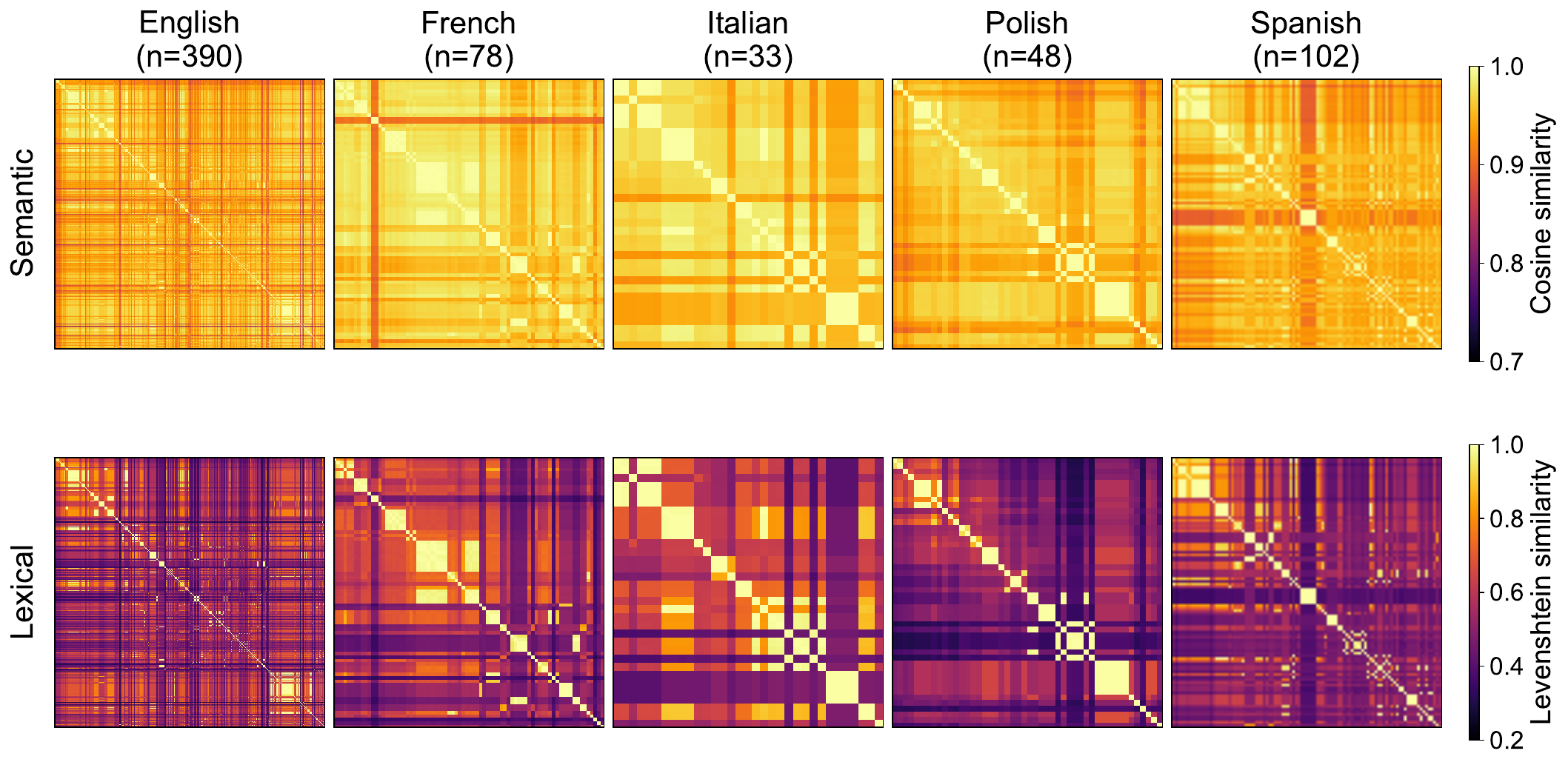}
    \caption{Pairwise similarity matrices between translations for each language, ordered by year of publication. Top row: semantic similarity (cosine on Qwen3-Embedding-8B chapter embeddings). Bottom row: lexical similarity (Levenshtein). Computed at the chapter level across all New Testament books.}
    \label{fig:similarity_matrix}
\end{figure*}

For the semantic comparison, we use the chapter-level Qwen3-Embedding-8B vectors distributed with the corpus (§\ref{sec:corpus_structure}). For each pair of translations sharing a chapter, we compute the cosine similarity of their embeddings and average across all shared chapters to produce a final score.

While effective for capturing broad stylistic differences, semantic similarity can fail to register minor textual variations. We therefore also use the pre-computed lexical similarity scores, based on the Levenshtein similarity defined as $1 - \frac{LD(x_1,x_2)}{\textrm{max}\left(\textrm{len}(x_1), \textrm{len}(x_2)\right)}$, where $LD(x_1, x_2)$ is the Levenshtein distance.

The pairwise similarity matrices in Figure \ref{fig:similarity_matrix} reveal the corpus's internal structure. From the semantic matrices, most translation pairs score above 0.8 in cosine similarity, reflecting their shared source text and narrative content.

Two types of clusters are visible. The most prominent are the tight, high-similarity blocks along the diagonal, representing near-identical editions or minor revisions of the same translation work. We also observe distinct clusters of translations that are internally cohesive but markedly dissimilar from the main group, often corresponding to dynamic or paraphrastic translations that employ a different stylistic register.

These clusters appear in both matrices, but their distinctiveness is far more pronounced in the lexical view. The lexical metric, sensitive to specific word choices and sentence structures, better captures the stylistic strategies of dynamic versions --- differences that the semantic embeddings partially smooth over.

\begin{figure}[htbp]
    \centering
    \includegraphics[width=\columnwidth]{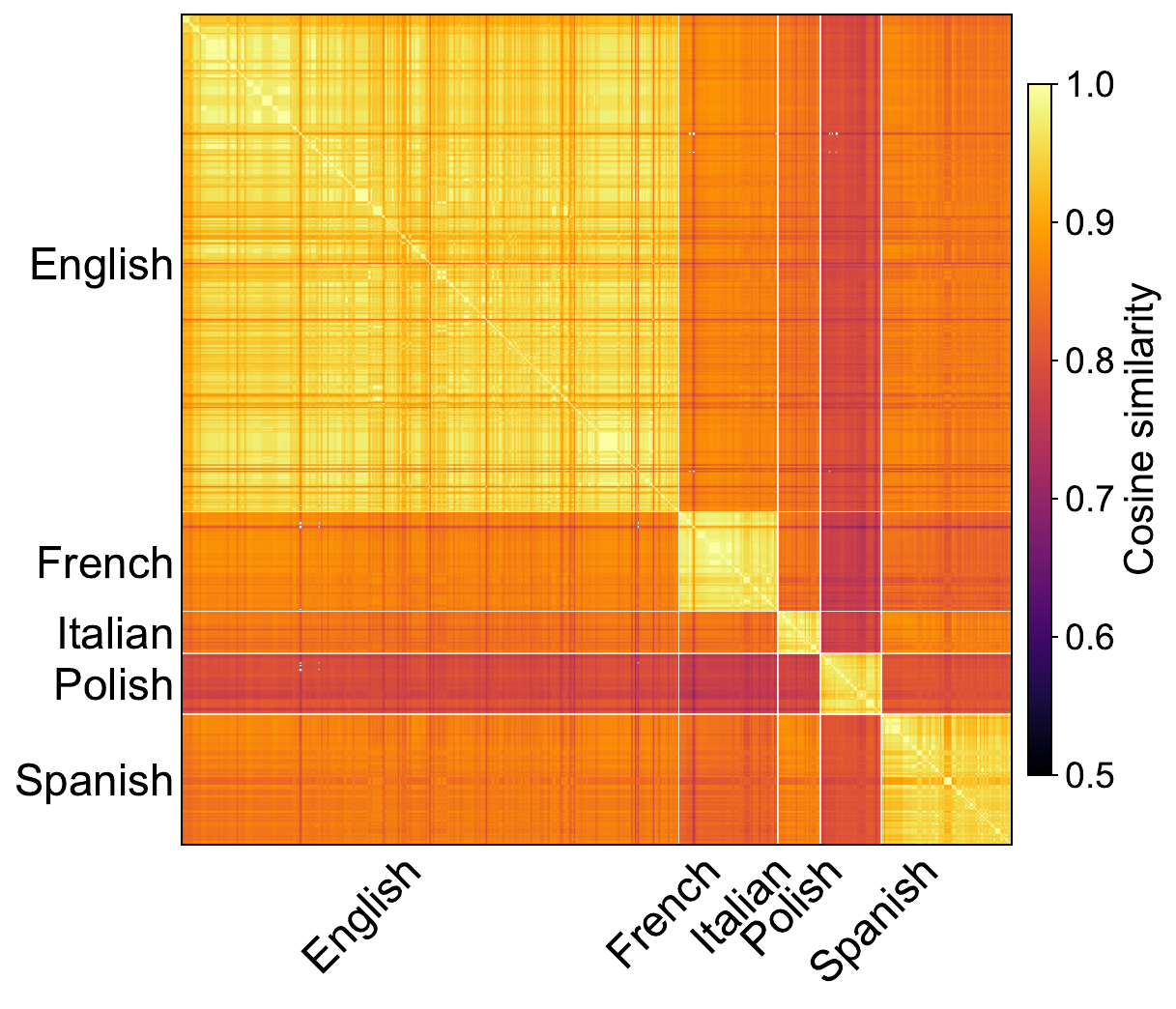}
    \caption{Cross-language semantic similarity matrix for all \data{total_translations} translations, sorted by language and year. Within-language blocks (along the diagonal) exhibit high similarity; off-diagonal blocks show cross-language semantic correspondence.}
    \label{fig:cross_language_similarity}
\end{figure}

Figure \ref{fig:cross_language_similarity} extends this analysis across all five languages. Within-language blocks along the diagonal exhibit the highest mean cosine similarity (\data{xsim_within_mean}), ranging from \data{xsim_within_min} (English, whose large and stylistically diverse collection lowers the average) to \data{xsim_within_max} (French and Italian). Cross-language similarity is lower overall (mean \data{xsim_cross_mean}), yet the block structure reveals a clear typological pattern. Among the Romance languages, Italian--Spanish reaches \data{xsim_italian_spanish_mean}, the highest cross-language pair, though English--French is comparable at \data{xsim_english_french_mean}. Polish, the sole Slavic language, is the most distant from every other language in the corpus: its cross-language means range from \data{xsim_french_polish_mean} (French--Polish, the lowest pair) to \data{xsim_polish_spanish_mean} (Polish--Spanish). English occupies an intermediate position, reflecting both its Germanic substrate and its Romance-derived vocabulary.

\subsection{Pairwise Variation}\label{sec:variation}

We use the chapter-level Levenshtein scores to validate our canonicalization and quantify inter-edition variation, distinguishing three tiers of relatedness: (a)~all within-language pairs, (b)~pairs sharing a \texttt{canonical\_id} (same translation family), and (c)~pairs sharing both \texttt{canonical\_id} and \texttt{canonical\_version} (same edition from different sources).

\begin{figure*}[htbp]
    \centering
    \includegraphics[width=\textwidth]{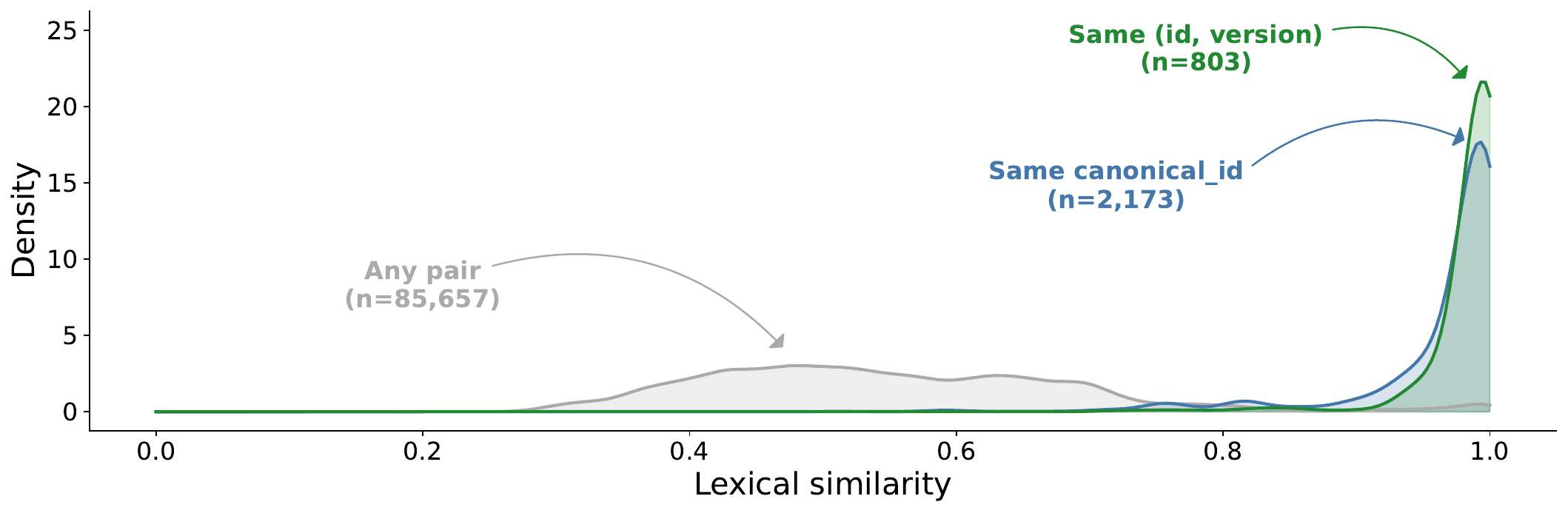}
    \caption{Lexical similarity distributions for three tiers of translation relatedness, pooled across all five languages: all within-language pairs (``Any pair''), pairs sharing a \texttt{canonical\_id}, and pairs sharing both \texttt{canonical\_id} and \texttt{canonical\_version}.}
    \label{fig:kde_variation}
\end{figure*}

Figure~\ref{fig:kde_variation} presents kernel density estimates for these three groups, pooled across all five languages. All within-language pairs center at a median of \data{sim_all_p50}. Same-\texttt{canonical\_id} pairs shift rightward (median \data{sim_cid_p50}), and same-version pairs cluster near 1.0 (median \data{sim_ver_p50}). Per-language percentiles (P10, P50, P90) for the two matched groups are reported in Appendix~\ref{sec:appendix_thresholds}. The per-language breakdown confirms that our canonicalization scheme reliably groups near-identical editions: same-version pairs achieve a P10 above 0.83 in every language, with median similarity at least 0.988. For same-\texttt{canonical\_id} pairs, the spread is wider (P10 ranging from 0.755 for Spanish to 0.986 for Polish), reflecting the varying degree of textual revision within translation families across languages.

Beyond validating the annotation, the pairwise scores allow us to test whether translation diversity has increased over time. For this analysis, we use the \data{unique_translations} unique translations (deduplicated by edition) to avoid pseudoreplication from cross-site copies. We regressed pairwise similarity on the midpoint year of each pair, controlling for temporal distance (both predictors standardized). The analysis covers all non-same-\texttt{canonical\_id} pairs within each language. A negative mid-year coefficient indicates that translations published in the same era are more diverse in recent centuries. The mid-year coefficient is significantly negative in every language for both lexical and semantic similarity (all $p \leq .011$): lexical $\beta$ ranges from $\data{sim_diachronic_english_lex_midyear_beta}$ (English) to $\data{sim_diachronic_spanish_lex_midyear_beta}$ (Spanish); semantic $\beta$ ranges from $\data{sim_diachronic_english_sem_midyear_beta}$ (English) to $\data{sim_diachronic_italian_sem_midyear_beta}$ (Italian). Full per-language regression results are reported in Appendix~\ref{sec:appendix_diachronic}. The effect sizes are modest ($R^2 = \data{sim_diachronic_r2_min}$--$\data{sim_diachronic_r2_max}$; see Appendix~\ref{sec:appendix_diachronic}), indicating that the temporal trend, while reliable, accounts for only a fraction of pairwise similarity variance --- most of which is driven by translation philosophy and other factors. Nevertheless, the consistent direction across all languages and both metrics indicates that modern translations are both lexically and semantically more diverse than historical ones, consistent with a broadening of translation strategies over time --- from narrower stylistic conventions in earlier centuries to the wide spectrum of approaches (literal, dynamic, paraphrase) seen today.




\section{Discussion}

\subsection{New Research Directions}

The \texttt{Targum} corpus makes tractable research questions that were previously difficult to address at scale. These avenues can be broadly grouped into \textit{micro-} and \textit{macrolevel} analyses.

\subsubsection{Micro-Level Analysis}

First, \texttt{Targum} facilitates \textit{micro-level analyses} by allowing researchers to select and compare focused subsets of the corpus based on specific metadata criteria. For example, scholars can:
\begin{itemize}
    \item \textbf{Conduct diachronic studies} by tracing linguistic changes across decades of revisions within a single translation family (e.g., the King James Version).
    \item \textbf{Perform synchronic comparisons} of contemporary versions to isolate the effects of confessional tradition or individual translator style.
    \item \textbf{Track translation lineage} by following the propagation of theological and stylistic choices through indirect translation chains, such as those mediated by a Latin Vulgate source or, in modern institutional cases (e.g., Jehovah's Witnesses, Biblica), by an English pivot translation.
\end{itemize}

Furthermore, the corpus can serve as a \textit{quantitative complement to traditional qualitative evaluation}. While an expert might label a translation as ``controversial'' or ``overly literal'' based on key passages, \texttt{Targum} allows researchers to measure the pervasiveness of these features across the entire New Testament, framing such judgments within the full scope of the text.

\subsubsection{Macro-Level Analysis}

Second, on a \textit{macro-level}, \texttt{Targum} can function as a \textit{benchmark} for any new translation. By generating a vector embedding for a new version, whether produced by humans or machines, it can be situated within the semantic space of the entire corpus. Its stylistic and theological profile can then be quantitatively compared against the full space of existing translations --- answering questions such as: ``Is this new translation stylistically closer to the KJV or the NIV family?'' or ``Does its interpretation of key theological terms align more with historical Catholic or Protestant traditions?''

\subsection{Future Work}

We plan to extend \texttt{Targum} along three axes: more translations, richer metadata, and broader language coverage.

\paragraph{More translations.}
We intend to add translations not yet captured by the current scraping scope, pursuing this along two lines. First, additional translations increase the historical representativeness of the corpus within each language, filling gaps in the NT translation record that the current source set leaves open. Second, adding further copies of translations already in the corpus --- sourced from sites not yet scraped --- functions as a cross-validation mechanism. Our site-specific parsing scripts are tested by comparing copies of the same translation that, by manual annotation, should be near-identical. When pairwise lexical similarity between such copies falls below the expected range, this signals a parsing error in one of the sources. Resolving these anomalies improves the accuracy of the affected site's parser, which in turn raises quality not only for translations with multiple copies but also for single-copy translations drawn from the same site, where analogous errors would otherwise go undetected.

\paragraph{Richer metadata.}
The current annotation covers canonical identity, revision year, and copyright status. We plan to add four fields. First, \textit{source text provenance} will record whether a translation was produced directly from the Greek, or mediated through the Latin Vulgate or an English intermediary --- a distinction that affects how closely a translation can be expected to track the Greek original. Second, \textit{translation lineage} will map dependency relationships between versions (e.g., which prior translations a given rendering explicitly acknowledges), enabling the modeling of complex genealogical flows that a flat canonical ID cannot represent. Third, \textit{confessional tradition} will classify each translation's institutional or theological affiliation. Fourth, \textit{critical edition} will identify the Greek source text used (e.g., Nestle-Aland 28th edition, Textus Receptus, Majority Text) where this can be established; this is feasible for most modern translations and less so for historical ones.

\paragraph{More languages.}
Extending coverage to other languages with rich NT translation traditions (e.g., German, Portuguese) is technically straightforward, but requires annotators with sufficient domain knowledge to verify canonical identities and assess translation history. We are open to expanding the language set, contingent on interest from the research community and the availability of collaborators who can contribute that expertise.

\section{Conclusion}

\texttt{Targum} addresses a key gap in existing biblical resources by providing a multilingual New Testament corpus focused on the depth of translation within each language. With extensive coverage for five European languages and \data{unique_translations} unique translations enriched with verified metadata, \texttt{Targum} provides the infrastructure for a new class of computational research --- addressing long-standing questions about the interplay of theology, style, and culture in translation using scalable quantitative methods.

\section{Limitations}

We acknowledge several limitations in the current version of the \texttt{Targum} corpus.

First, the corpus is necessarily constrained by the availability of digitized translations on public websites. This introduces a potential ``digitization bias'': the corpus reflects the history of translations that have been made digitally accessible, which may not perfectly align with the complete publication history. For instance, German was initially considered for inclusion due to its rich translation tradition but was ultimately excluded because of the relative scarcity of online versions compared to the five languages that were selected. Therefore, the distribution of translations in \texttt{Targum} should be understood as representative of the digital landscape, not the complete landscape.

Second, the manual normalization process has inherent limitations. Although our methodology is systematic, the final classification is still subject to human error. A key challenge arose when trying to pinpoint the exact revision year for certain translations. This difficulty was most pronounced not with historical texts, which often have well-documented print editions, but with modern, digital-native translations distributed primarily online. Such versions can be revised multiple times per year without formal versioning, and their release history is often undocumented. In these cases, we made a best-effort judgment based on the available textual and contextual evidence (e.g., copyright dates on the webpage) to determine the edition. A complementary difficulty arises at the opposite end of the time axis: historical translations that have undergone modernized re-editions. For example, Wycliffe's Bible received a modernized-spelling edition for its 500th anniversary, and many other historical translations have been periodically revised over the centuries. In such cases the translation philosophy is historical while the orthography is contemporary, making it difficult to assign a single revision year that captures both dimensions. When the time span between the original and the revision is large the distinction is clear, but for translations with recurrent, incremental updates the boundary is less obvious.

\section{Ethics Statement}

We aggregated data for the \texttt{Targum} corpus from publicly accessible websites. The authors do not claim copyright on any of the translated texts. As shown in Table \ref{tab:copyright_stats}, the copyright status varies across languages. Our distribution plan respects the intellectual property rights of the copyright holders while maximising the utility of the resource for the research community. Translations in the public domain or behind open licenses, along with the pre-computed embeddings and pairwise similarity scores for all translations, will be made fully publicly available (at \url{https://github.com/mrapacz/targum-corpus}). Copyrighted texts will not be publicly distributed in full; instead, they will be made available upon reasonable request for noncommercial research purposes. We will remove any translation from the corpus upon a direct request from a verified copyright holder.

In the preparation of this work, AI assistants were used as productivity tools. During initial development, Gemini 2.5 Pro (via Cursor and the Gemini web interface) was used for tasks such as generating boilerplate code for web scraping and helping with the editing and formatting of the submitted manuscript. For the camera-ready version, Claude Opus 4.6 and Sonnet 4.6 (via GitHub Copilot) were used for agentic engineering tasks such as generating code for data analysis and validation of data prior to publishing, and helping with the editing and formatting of this article.

\section{Acknowledgments}
This research was supported by the National Science Centre, Poland, under project number 2025/57/N/HS2/04961.
We gratefully acknowledge the Polish high-performance computing infrastructure PLGrid (HPC Center: ACK Cyfronet AGH) for providing computational facilities and support under grant no. PLG/2026/019145.

\section{Bibliographical References}\label{sec:reference}

\renewcommand{\bibsection}{} 
\bibliographystyle{lrec2026-natbib}
\bibliography{bibliography}

\bibliographystylelanguageresource{lrec2026-natbib}
\bibliographylanguageresource{languageresource}

\onecolumn
\appendix

\section{Corpus Metadata Fields}\label{sec:appendix_metadata}

Each translation in the \texttt{Targum} corpus is described by the metadata fields listed in Table~\ref{tab:metadata_fields}. These fields are derived from the source websites and our manual annotation process.

\begin{table}[h]
    \centering
    \begin{tabular}{@{} l p{0.45\linewidth} l @{}}
        \toprule
        Field & Description & Example \\
        \midrule
        \texttt{site} & Source website from which the translation was acquired. & \texttt{laparola.net} \\
        \texttt{iso} & ISO 639-3 language code. & \texttt{eng} \\
        \texttt{translation\_id} & Site-specific identifier assigned by the source. & \texttt{asv} \\
        \texttt{translation\_name} & Full human-readable name. & \texttt{American Standard Version} \\
        \texttt{translation\_abbr} & Common abbreviation. & \texttt{ASV} \\
        \texttt{canonical\_id} & Groups all instances of the same translation work across sites and editions. & \texttt{king-james-version} \\
        \texttt{canonical\_version} & Edition label; often a year but may encode richer distinctions (e.g.\ \texttt{1989-anglicised-catholic}). & \texttt{1769} \\
        \texttt{canonical\_year} & Integer revision year; places the edition on the time axis. & \texttt{1769} \\
        \texttt{num\_chapters} & Number of New Testament chapters present. & \texttt{260} \\
        \texttt{num\_verses} & Number of New Testament verses present. & \texttt{7957} \\
        \texttt{num\_words} & Total word count across all chapters. & \texttt{179836} \\
        \texttt{copyright\_status} & Copyright classification; accompanied by a reasoning field (not shown). & \texttt{public\_domain} \\
        \bottomrule
    \end{tabular}
    \caption{Metadata fields available for each translation in the \texttt{Targum} corpus.}
    \label{tab:metadata_fields}
\end{table}

\section{Corpus File Hierarchy}\label{sec:appendix_hierarchy}

The corpus repository has the following top-level structure:

\begin{small}
\begin{verbatim}
targum-corpus/
  index.tsv
  copyrights.tsv
  book_coverage.tsv
  manifest.json
  corpus/
    {site}/
      {iso}/
        {id}.jsonl
  similarity/
    lexical/
      levenshtein/chapter/{iso}.parquet
    semantic/
      cosine/
        {model}/chapter/{iso}.parquet
        {model}/chapter/cross_language.parquet
  embeddings/
    {model}/
      language={iso}/
        site={site}/
          translation={id}/
            granularity={chapter|verse}/
              data.parquet
\end{verbatim}
\end{small}

\noindent
The placeholders expand as follows: \texttt{\{site\}} is the source library (e.g.\ \texttt{ebible.org}, \texttt{biblegateway.com}); \texttt{\{iso\}} is the ISO~639-3 language code (\texttt{eng}, \texttt{fra}, \texttt{ita}, \texttt{pol}, \texttt{spa}); \texttt{\{id\}} is the source-specific translation identifier listed in \texttt{index.tsv}; and \texttt{\{model\}} encodes the embedding model with the separator \texttt{XxX} (e.g.\ \texttt{QwenXxXQwen3-Embedding-8B}).

\paragraph{Root-level metadata.}
Four files at the repository root describe the corpus as a whole.
\texttt{index.tsv} lists every translation instance with the metadata fields described in Appendix~\ref{sec:appendix_metadata}.
\texttt{copyrights.tsv} records the copyright text and status for each translation.
\texttt{book\_coverage.tsv} indicates which of the 27 New Testament books each translation contains.
\texttt{manifest.json} provides summary statistics (total counts, per-language breakdowns).

\paragraph{Translation texts (\texttt{corpus/}).}
Each translation is stored as a JSONL file (one JSON object per verse) under \texttt{corpus/\{site\}/\{iso\}/}. Each line contains four fields, as illustrated below with the first three verses of the King James Version (Matthew~1:1--3):

\begin{small}
\begin{verbatim}
{"book":"MAT","chapter":1,"verse":"1",
 "text":"The book of the generation of
 Jesus Christ, the son of David, the
 son of Abraham."}
{"book":"MAT","chapter":1,"verse":"2",
 "text":"Abraham begat Isaac; and Isaac
 begat Jacob; and Jacob begat Judas and
 his brethren;"}
{"book":"MAT","chapter":1,"verse":"3",
 "text":"and Judas begat Phares and Zara
 of Thamar; and Phares begat Esrom; and
 Esrom begat Aram;"}
\end{verbatim}
\end{small}

\noindent
The \texttt{book} field uses USFM three-letter codes (e.g.\ \texttt{MAT}, \texttt{MRK}, \texttt{REV}); \texttt{chapter} is an integer; \texttt{verse} is a string to accommodate non-standard segmentation (e.g.\ \texttt{"6a"}, \texttt{"6-8"}).

\paragraph{Pairwise similarity scores (\texttt{similarity/}).}
Pre-computed chapter-level pairwise similarity scores are stored as Parquet files, one per language. Two metrics are provided: Levenshtein (lexical) and cosine distance over Qwen3-Embedding-8B vectors (semantic). The semantic directory also contains a \texttt{cross\_language.parquet} file with all inter-lingual pairs.

\paragraph{Text embeddings (\texttt{embeddings/}).}
Embeddings are stored as Hive-partitioned Parquet files. Two models are included (Qwen3-Embedding-0.6B and Qwen3-Embedding-8B), each at chapter and verse granularity. Each \texttt{data.parquet} file contains two columns: \texttt{key} (e.g.\ \texttt{MAT~1} for chapter, \texttt{MAT~1:1} for verse) and \texttt{value} (the embedding vector).

\section{Similarity Thresholds}\label{sec:appendix_thresholds}

Table~\ref{tab:sim_thresholds} reports per-language lexical similarity percentiles for translation pairs at three levels of relatedness: any within-language pair, pairs sharing a canonical identity, and pairs sharing both canonical identity and version.

\begin{table}[h]
    \centering
    \begin{tabular}{l*{3}{r}*{3}{r}*{3}{r}}
    \toprule
     & \multicolumn{3}{c}{Any pair} & \multicolumn{3}{c}{Same \texttt{canonical\_id}} & \multicolumn{3}{c}{Same (id, version)} \\
    \cmidrule(lr){2-4} \cmidrule(lr){5-7} \cmidrule(lr){8-10}
    Language & P10 & P50 & P90 & P10 & P50 & P90 & P10 & P50 & P90 \\
    \midrule
    English & 0.386 & 0.535 & 0.710 & 0.944 & 0.991 & 1.000 & 0.963 & 0.995 & 1.000 \\
    French & 0.419 & 0.552 & 0.732 & 0.903 & 0.986 & 1.000 & 0.963 & 0.988 & 1.000 \\
    Italian & 0.423 & 0.601 & 0.893 & 0.835 & 0.991 & 1.000 & 0.835 & 0.998 & 1.000 \\
    Polish & 0.355 & 0.492 & 0.648 & 0.986 & 0.998 & 1.000 & 0.993 & 0.999 & 1.000 \\
    Spanish & 0.411 & 0.491 & 0.750 & 0.755 & 0.956 & 1.000 & 0.990 & 0.999 & 1.000 \\
    \bottomrule
\end{tabular}
    \caption{Lexical similarity percentiles (P10, P50, P90) for within-language translation pairs: any pair, pairs sharing a \texttt{canonical\_id}, and pairs sharing both \texttt{canonical\_id} and \texttt{canonical\_version}.}
    \label{tab:sim_thresholds}
\end{table}

\section{Diachronic Diversity Regressions}\label{sec:appendix_diachronic}

Table~\ref{tab:diachronic} reports the OLS regression results for the diachronic diversity analysis. For each language, we regressed pairwise similarity (lexical and semantic) on the midpoint year of the pair, controlling for temporal distance between the two translations (both predictors standardized). Only non-same-\texttt{canonical\_id} pairs among the \data{unique_translations} unique translations (deduplicated by edition) are included. A negative mid-year coefficient ($\beta$) indicates that translations produced in the same era are more diverse in more recent centuries.

\begin{table}[h]
    \centering
    \begin{tabular}{l rrrr rrrr}
    \toprule
     & \multicolumn{4}{c}{Lexical} & \multicolumn{4}{c}{Semantic} \\
    \cmidrule(lr){2-5} \cmidrule(lr){6-9}
    Language & $N$ & $\beta$ & $p$ & $R^2$ & $N$ & $\beta$ & $p$ & $R^2$ \\
    \midrule
    English & 18,528 & $-0.03402$ & $< .001$ & 0.0197 & 18,528 & $-0.00340$ & $< .001$ & 0.0537 \\
    French & 814 & $-0.05848$ & $< .001$ & 0.1341 & 814 & $-0.00431$ & $< .001$ & 0.0297 \\
    Italian & 129 & $-0.05390$ & $= 0.011$ & 0.0585 & 129 & $-0.00987$ & $< .001$ & 0.1198 \\
    Polish & 402 & $-0.07290$ & $< .001$ & 0.2367 & 402 & $-0.00390$ & $= 0.001$ & 0.2790 \\
    Spanish & 1,348 & $-0.10069$ & $< .001$ & 0.1169 & 1,348 & $-0.00805$ & $< .001$ & 0.0915 \\
    \bottomrule
    \multicolumn{9}{l}{\small $\beta$ = standardized mid-year coefficient; negative values indicate increasing diversity over time.} \\
\end{tabular}
    \caption{OLS regression of pairwise similarity on mid-year for non-same-\texttt{canonical\_id} pairs, per language. Negative $\beta$ indicates increasing diversity over time.}
    \label{tab:diachronic}
\end{table}

\end{document}